\documentclass{article}

\PassOptionsToPackage{numbers, compress}{natbib}

\usepackage[eandd, final]{neurips_2026}

\usepackage{booktabs}
\usepackage{threeparttable}   
\usepackage{pifont}
\usepackage[utf8]{inputenc}
\usepackage[T1]{fontenc}
\usepackage{hyperref}
\usepackage{float}
\usepackage{url}
\usepackage{booktabs}
\usepackage{amsfonts}
\usepackage{amsmath}
\usepackage{nicefrac}
\usepackage{microtype}
\usepackage{graphicx}
\usepackage{xcolor}
\usepackage{multirow}
\usepackage{authblk}
\usepackage{makecell}
\usepackage{colortbl}
\usepackage{enumitem}

\newcommand{\cmark}{\ding{51}} 
\newcommand{\xmark}{\ding{55}} 
\newcommand{\pmark}{\textcolor{red}{\ensuremath{\circ}}}   
\newcommand{\gcmark}{\textcolor{green!80!black}{\ding{51}}}
\newcommand{\rxmark}{\textcolor{red}{\ding{55}}}

\title{OpenWhistle: A Large-Scale Longitudinal Dataset and Benchmark
of Bottlenose Dolphin Vocalizations}

\author{
  \textbf{Faadil Mustun}$^{1,*}$ \quad
  \textbf{Chiara Semenzin}$^{2,*,\dagger}$ \quad
  \textbf{Roberto Dessì}$^{3}$ \quad
  \textbf{Pablo Robin Guerrero}$^{1}$ \quad
  \textbf{Pierre Orhan}$^{4}$ \quad
  \textbf{Alexis Emanuelli}$^{1}$ \quad
  \textbf{Emanuele Rossi}$^{5}$ \quad
  \textbf{Yair Lakretz}$^{6}$ \quad
  \textbf{Gonzalo de~Polavieja}$^{7}$ \quad
  \textbf{Germán Sumbre}$^{1}$ \\
  \vspace{3pt}
  {\small\parbox{\linewidth}{\centering
  $^{1}$Institut de Biologie de l'École normale supérieure, CNRS, INSERM, Université PSL, Paris, France \quad
  $^{2}$Earth Species Project, France \quad
  $^{3}$Not Diamond, San Francisco, USA \quad
  $^{4}$Institut du Cerveau, Paris, France \quad
  $^{5}$Sapienza University of Rome, Rome, Italy \quad
  $^{6}$École normale supérieure, PSL University, EHESS, LSCP, CNRS. \quad
  $^{7}$Champalimaud Foundation, Lisbon, Portugal}}
  \\
  \vspace{3pt}
  {\small\parbox{\linewidth}{\centering
  $^{*}$These authors contributed equally to this work. \quad
  $^{\dagger}$Work carried out while at the Institut de Biologie de l'École normale supérieure, Paris, France.}}
}

\begin{document}

\maketitle

\begin{abstract}
    
    Recent advances in bioacoustics have been driven by large-scale corpora and standardized benchmarks, yet existing resources are overwhelmingly bird-centric and shallow per species, limiting their use for studying the structure of a single species' communication system. This gap is particularly acute for cetaceans: despite bottlenose dolphins (\textit{Tursiops truncatus}) being a compelling case of complex vocal communication among non-human mammals, existing dolphin datasets are small, fragmented, and largely closed. We introduce OpenWhistle, the largest publicly available dataset of dolphin vocalizations. It  comprises approximately 180,000 whistles (114 hours) recorded over five years from a stable pod of five individuals in a semi-natural environment, paired with a curated subset of 8,354 expert-annotated whistles and reproducible evaluation protocols for whistle-type detection and classification. We further release the full processing pipeline for whistle detection, segmentation, and categorization. To demonstrate its utility, we pretrain wav2vec 2.0 and AVES models on the OpenWhistle corpus and show that in-domain pretraining yields effective representations, outperforming off-the-shelf pretrained models, including the original AVES and BioLingual, on both tasks while leaving meaningful headroom for future work. By releasing the dataset, pipeline, and evaluation protocol, we provide the first open dolphin whistle dataset tailored for training self-supervised models, laying the groundwork for advancing dolphin communication research and developing models that capture fine-grained acoustic structure within species. The dataset, models, and code are available at \url{https://huggingface.co/collections/dolphinteam/neurips26-openwhistle} and \url{https://github.com/sumbre-lab/OpenWhistle}.
    
\end{abstract}

\section{Introduction}

Bioacoustics plays a central role in ecology and conservation, enabling researchers to study animal communication, monitor biodiversity, and track endangered species through acoustic signals~\citep{bradbury_principles_1998, laiolo2010emerging, fischer2013bioacoustic, rutz2023using}. 
The field has recently seen major advances in tasks such as detection, classification~\citep{stowell2022computational} and denoising~\citep{miron2024biodenoisinganimalvocalizationdenoising}, driven by machine-learning models~\citep{robinson2023transferablemodelsbioacousticshuman, robinson2025naturelm, vanmerrienboer2025perch20bitternlesson} and enabled by large-scale pretraining corpora, including Xeno-Canto~\citep{vellinga2015xeno}, iNaturalist~\citep{inaturalist_platform}, and Animal Sound Archive~\citep{animal_sound_archive_berlin}, together with standardized benchmarks such as BEANS~\citep{hagiwara2023beans}, BEANS-ZERO~\citep{robinson2025naturelm}, and BirdSet~\citep{rauch2024birdset}.

However, these corpora are broad in taxonomic coverage, but shallow for any single species: they aggregate short recordings across thousands of species, which suits detection and species classification but is insufficient for studying the structure of a species' communication system. Questions about vocal learning, individual identity, social coordination, and temporal change require deep, longitudinal data from known individuals of a single species, a resource that to the best of our knowledge does not exist at scale.  Bottlenose dolphins (\textit{Tursiops truncatus}) represent one of the most compelling cases of complex vocal communication among non-human mammals, with individually distinctive signature whistles and documented vocal learning~\citep{janik2013communication}, making them one of the species for which such a resource would be most valuable. Despite extensive study, progress toward understanding dolphin communication using AI tools has been limited by the lack of suitable data: existing dolphin whistle datasets are small, fragmented, and largely not publicly available.

We address this gap by introducing OpenWhistle, an open resource for dolphin vocalization research comprising two components: (i) a large-scale training corpus of approximately 180,000 dolphin whistles (114 hours) collected over five years from a pod of five known individuals in a semi-natural marine environment, and (ii) a curated dataset with 8,354 expert-annotated labels and explicit evaluation protocols for two tasks: whistle detection and whistle-type classification. Beyond these core tasks, OpenWhistle was designed to preserve contiguous whistle sequences from interacting individuals across five years, enabling future work on richer biological questions such as individual variation, vocal exchanges, interaction dynamics, temporal drift, and vocal development.

Beyond its scientific value, OpenWhistle complements broad-coverage bioacoustic datasets and benchmarks~\citep{hagiwara2023beans, rauch2024birdset} by providing a deep, longitudinal corpus from a single communication system, with known individuals and expert whistle-type labels.
To our knowledge, it is the first open, ML-ready single-species cetacean dataset of sufficient scale for self-supervised pretraining directly from raw audio, enabling direct comparison between in-domain specialization and broad-coverage pretraining for fine-grained acoustic discrimination.
Its pairing of a large unlabeled corpus with a smaller expert-annotated subset also makes it a natural testbed for label-efficient methods such as semi-supervised, active, and few-shot learning, addressing a bottleneck repeatedly identified in bioacoustics~\citep{stowell2022computational, rutz2023using, Hagiwara:etal:2022, schaeferzimmermann2026animal2vec, nolasco2022fewshot, mcewen2024activefewshot}.
Finally, its continuous recordings preserve environmental sounds, variable SNR, and overlapping vocalizations~\citep{miron2024biodenoisinganimalvocalizationdenoising}, while its longitudinal structure supports temporal distribution shift and continual-learning evaluations within a single known-individual population, complementing broader covariate-shift benchmarks such as BirdSet~\citep{rauch2024birdset}.

Our contributions are as follows:
\begin{itemize}[leftmargin=1.2em]
\item \textbf{OpenWhistle dataset:} We release\footnote{\url{https://huggingface.co/collections/dolphinteam/neurips26-openwhistle}} the largest publicly available dataset of dolphin vocalizations to date, with three key properties:
\begin{itemize}[label=$\circ$, leftmargin=1.2em, topsep=2pt, itemsep=1pt]
    \item \textbf{Scale:} around 180,000 whistles (114 hours) from a stable pod of known individuals.
    \item \textbf{Expert annotations and benchmark:} a curated subset of 8,354 expert-annotated whistles with reproducible evaluation protocols for whistle-type detection and classification.
    \item \textbf{Longitudinal structure:} contiguous whistle sequences spanning five years, enabling future work on vocal exchanges, interaction dynamics, and temporal drift.
\end{itemize}
\item \textbf{Annotation pipeline:} We release\footnote{\url{https://github.com/sumbre-lab/OpenWhistle}} a scalable pipeline for whistle detection, segmentation, and type categorization, offering a practical recipe for constructing large dolphin acoustic datasets from continuous passive acoustic monitoring recordings.
\item \textbf{In-domain pretraining baselines:} We show that self-supervised models pretrained on OpenWhistle with two architectures, wav2vec 2.0~\citep{Baevski:etal:2020} and AVES-bio~\citep{Hagiwara:etal:2022}, learn effective representations of dolphin whistles, outperforming off-the-shelf pretrained models on both benchmark tasks and establishing that in-domain data provides a meaningful advantage for both architectures.
\end{itemize}

\section{Related Work}

\subsection{Dolphin Vocalizations and Communication}
\label{sec:related_work_dolphin}

Early work by~\cite{connor1996pop} and~\cite{janik2013communication} established that dolphin communication relies primarily on two types of sounds: burst pulses and whistles, with whistles playing a central role in social interactions. Among whistles, signature whistles (SW) were shown by~\cite{sayigh2007facts} and~\cite{janik2000whistle} to be stable, individually distinctive calls used for recognition and maintaining social bonds. These studies demonstrated that dolphins develop unique acoustic identifiers and can both produce their own signature whistle and imitate those of conspecifics. \cite{sayigh2007facts} found that signature whistles dominate dolphin vocal repertoires, making up as much as 70\% of whistles recorded in natural settings. Non-signature whistles (NSW), which comprise the remainder of the whistle repertoire, are more variable in structure and are not uniquely associated with individuals. Their communicative role remains less well understood~\citep{janik2014cetacean}.

\subsection{Existing Datasets}
\label{sec:related_work_datasets}

\begin{table}[h]
    \renewcommand{\arraystretch}{1.2}
    \centering
    \resizebox{\columnwidth}{!}{%
    \begin{tabular}{l r r c c c c c}
    \toprule
    Dataset 
    & \# Whistles 
    & \makecell{Voc.\\hours}
    & \makecell{Time\\span\\(yrs)}
    & \makecell{Stable pod\\(\# indiv.)}
    & Setting
    & \makecell{Seq.\\context}
    & Open \\
    \midrule
    
    \textbf{OpenWhistle Pretraining}
    & \textbf{$\sim$180,000*}
    & \textbf{114.3}
    & \textbf{5.0}
    & \textbf{\cmark{} (5)}
    & \textbf{Semi-natural}
    & \textbf{\cmark}
    & \textbf{\gcmark} \\
    
    \textbf{OpenWhistle Expert subset}
    & \textbf{8,354}
    & \textbf{1.9}
    & \textbf{0.42}
    & \textbf{\cmark{} (5)}
    & \textbf{Semi-natural}
    & \textbf{\xmark}
    & \textbf{\gcmark} \\

    DOLPHINFREE~\cite{lehnhoff2025}
    & 4,600
    & 7.3
    & 2.0
    & \xmark
    & Wild
    & \xmark
    & \gcmark \\
    
    Di Nardo et al., 2025~\cite{dinardo2025}
    & 3,111
    & 0.6
    & 0.003
    & \cmark{} (7)
    & Captive
    & \xmark
    & \gcmark \\
    
    Watkins MMSD~\cite{sayigh2016watkins}
    & 566
    & N/R
    & 70+
    & \xmark
    & Wild
    & \xmark
    & \gcmark \\
    
    \midrule
    
    Korkmaz et al., 2023~\cite{korkmaz2023}
    & $\sim$29,000*
    & 6.8
    & 0.07
    & \xmark
    & Semi-natural
    & \cmark
    & \pmark \\
    
    Sicily Strait PAM~\cite{gregorietti2021}
    & 14,048
    & N/R
    & 1.2
    & \xmark
    & Wild
    & \cmark
    & \rxmark \\
    
    DCLDE 2011~\cite{roch2025dclde}
    & 6,011
    & 0.7
    & 4.0
    & \xmark
    & Wild
    & \xmark
    & \rxmark \\
    
    SDWD~\cite{sayigh2022sarasota}
    & N/R
    & N/R
    & 43+
    & \cmark{} (293)
    & Wild (\textit{Catch-\&-Rel.})
    & \xmark
    & \pmark \\
    
    \bottomrule
    \end{tabular}
    }
    \caption{Comparison of existing dolphin acoustic datasets.
    N/R = not reported;
    \pmark = available upon request.
    Time span is reported in years for consistency across datasets.
    "Seq. context" indicates whether the dataset preserves temporally contiguous sequences of multiple whistles, rather than only isolated whistle clips.
    All datasets are based on passive acoustic monitoring (PAM), except SDWD, which includes data collected through catch-and-release protocols.
    * Estimated from total vocalization duration and mean whistle duration.
    }
    \label{tab:datasets}
\end{table}

Large-scale bioacoustic datasets have played a central role in recent progress in the field, but they are overwhelmingly bird-centric, with resources such as Xeno-Canto and BirdSet dominating the landscape~\cite{vellinga2015xeno,rauch2024birdset}. These datasets provide broad taxonomic coverage and large volumes of data, but are typically shallow per species and focus on detection or species classification, making them less suitable for studying the structure of the communication system of a given species.

In contrast, dolphin acoustic datasets remain limited in both scale and accessibility (Table~\ref{tab:datasets}). Existing resources fall into three main categories. First, small, high-quality datasets such as DCLDE 2011~\cite{li2023,roch2025dclde} and DOLPHINFREE~\cite{lehnhoff2025} provide detailed contour annotations, but contain only a few thousand whistles, limiting their use for data-intensive methods. Similarly, Di Nardo et al.~\cite{dinardo2025} provide curated whistle data, but at a smaller scale and in a captive environment. Unlike OpenWhistle, they lack the scale required for data-intensive methods such as self-supervised learning. Second, passive acoustic monitoring datasets, such as the Sicily Strait recordings~\cite{gregorietti2021}, offer longer temporal coverage in wild settings but typically lack fine-grained annotations, often reporting only the presence of vocal activity. In contrast, our dataset provides whistle-level labels together with continuous recordings from known individuals. Finally, specialized datasets such as SDWD~\cite{sayigh2022sarasota} focus on specific aspects like individual identity, but are not fully open for general use or large-scale machine learning. 

More recent efforts, such as \cite{korkmaz2023}, increase dataset size but introduce other constraints, including the use of spectrogram images instead of raw audio and coarse binary annotations. In contrast, OpenWhistle provides raw audio, fine-grained whistle-type annotations, and a reproducible evaluation protocol. 
No existing resource combines large-scale, open-access, longitudinal recordings from known individuals with well-documented histories and whistle-level annotations, gaps that OpenWhistle is designed to fill. It is also the only such resource tested for self-supervised models.

\section{Data Collection}
\label{sec:data_collection}

\begin{figure}[t]
    \centering
    \includegraphics[width=1\textwidth]{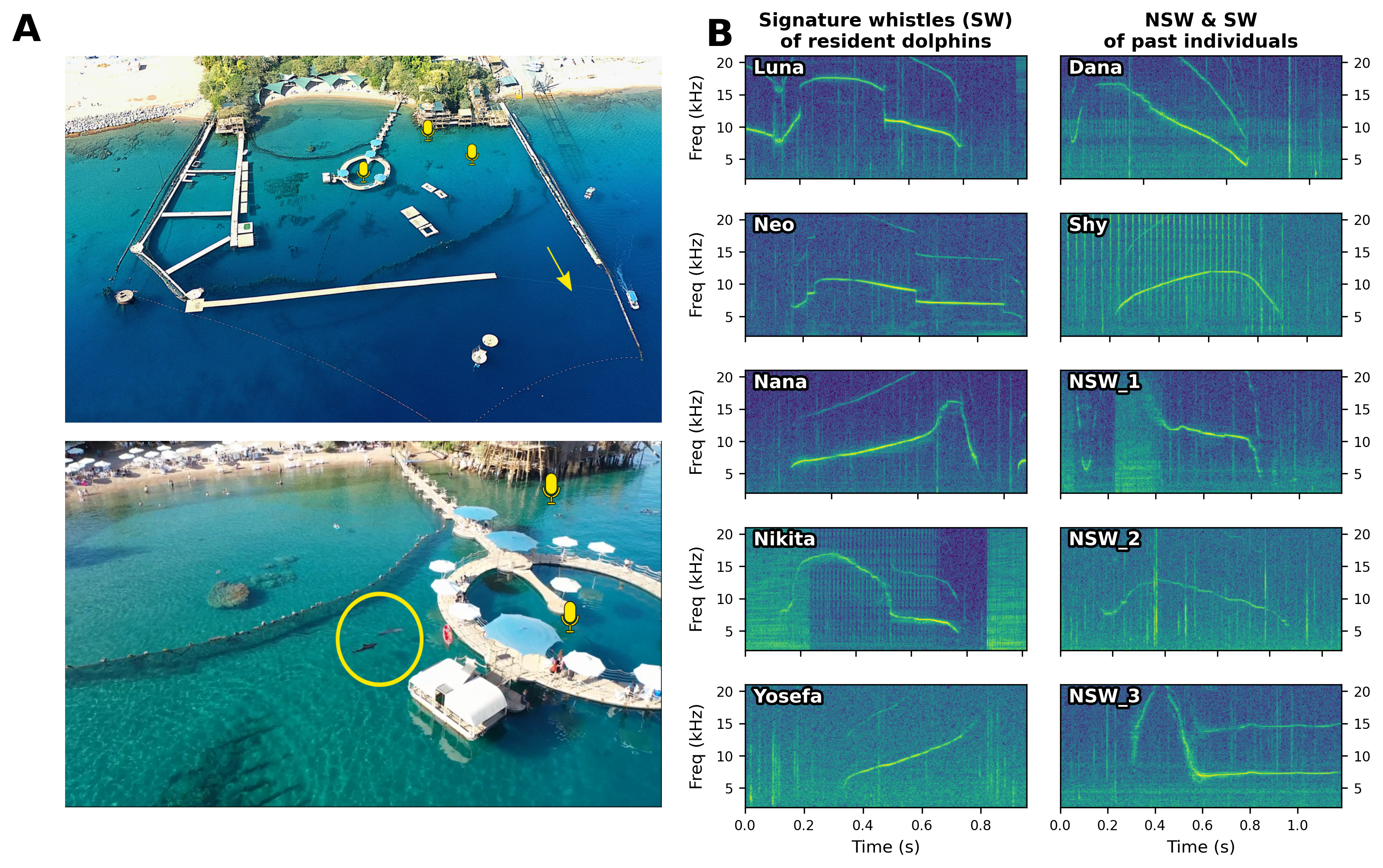}
    \caption{\textbf{Site Description and Whistle Repertoire.} A) The unique recording site at Dolphin Reef, Eilat. Hydrophones (yellow markers) are deployed at fixed locations to continuously capture underwater audio. Dolphins move freely within the area and can exit to the open sea. B) Representative spectrograms of whistle types. Left: Signature Whistles (SW) of resident dolphins, each showing individually distinctive frequency contours. Right: Whistles of past individuals and non-signature whistles (NSW), illustrating the diversity of vocalizations captured in the dataset.}
    \label{fig:data_acquisition}
\end{figure}


Recordings were collected at Dolphin Reef, a coastal site on the northern Gulf of Aqaba. The site hosts a resident pod of \textit{Tursiops truncatus ponticus} in a large, delimited natural marine area open to the sea, enabling semi-natural behaviour while supporting long-term and continuous tracking of known individuals~\citep{perelberg2010studying}. Human-dolphin interactions occur only when initiated by the dolphins and are entirely voluntary. The dataset includes vocalizations from five dolphins: one male and three females, and one \textit{Tursiops aduncus} female from the Indian Ocean, who joined the pod sporadically in 2019. Dolphins tend to remain near the monitored area during periods of human presence, but frequently leave to forage in the open sea when the site is closed or human activity is low.

This setting has the advantage of both controlled captive studies and fully wild passive acoustic monitoring. Unlike captive environments, it preserves ecologically valid behaviour and realistic acoustic conditions, including natural social interactions. At the same time, unlike wild recordings, it provides stable individual identity, longitudinal continuity, and contextual interpretability over multiple years. This combination enables analyses that require both ecological realism and individual-level resolution, which are typically difficult to achieve simultaneously in bioacoustic datasets~\citep{perelberg2010studying}.

\begin{figure}[t]
    \centering
    \includegraphics[width=1\textwidth]{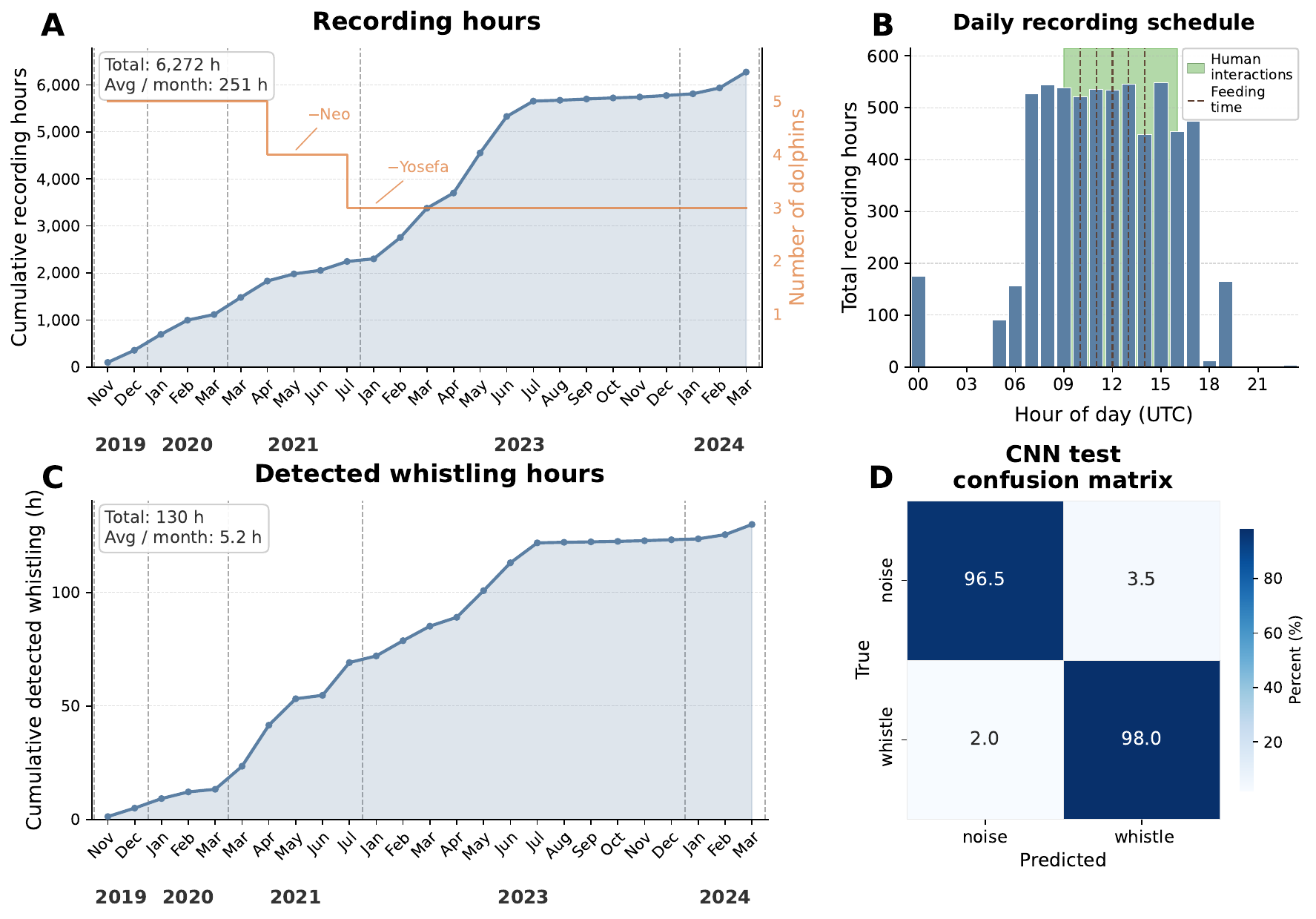}
    \caption{\textbf{OpenWhistle: Longitudinal Extent and Temporal Distribution of the Dataset.} A) Cumulative recording hours over time, showing dataset growth and changes in pod composition. 
    B) Distribution of recording hours across the day, indicating alignment with periods of human activity.
    C) Cumulative detected whistling hours over time, obtained by applying the whistle presence detection CNN to the raw recordings.
    D) Confusion matrix of the whistle presence detection CNN on the test set, indicating high reliability of the detected whistle segments used to derive panel C.}
    \label{fig:long_data}
\end{figure}

\section{Dataset Construction and Annotation}
\label{sec:datasets_construction}

\subsection{Annotation Pipeline}
\label{sec:pipeline}

\paragraph{Binary Whistle Presence Detection}
Raw audio was processed using a convolutional neural network based on the VGG16 architecture~\citep{simonyan2015vgg}, using ImageNet-pretrained weights~\citep{deng2009imagenet} and fine-tuned on a balanced 59,808-segment dataset (whistle vs non-whistle). The network takes spectrograms as input and outputs binary predictions indicating the presence of at least one whistle. On a held-out test set of 16,708 segments, the model achieved a precision of 96.52\% and a recall of 97.99\% (Figure~\ref{fig:long_data}D).

\paragraph{Whistle Segmentation}
The CNN operates on non-overlapping 0.4\,s windows. Consecutive detections were concatenated into continuous segments. To capture temporal structure, segments separated by less than 6\,s were merged into the same sequence, yielding variable-length whistle sequences.

\paragraph{Whistle Annotation} 
For the expert-annotated subset, detected whistle segments were categorized using ARTwarp~\citep{deecke2006automated}, an unsupervised neural network algorithm incorporating dynamic time warping (DTW)~\citep{buck1993quantitative} to cluster whistles by contour similarity. Following the procedure in~\citep{mustun2024whistle}, each whistle was assigned to one of 10 known categories by comparison with manually annotated template contours~\citep{roch_automated_2011}. The resulting assignments were manually refined through visual inspection of spectrograms by expert annotators, correcting misclassifications and resolving ambiguous cases. This two-stage procedure combines scalable unsupervised clustering with expert validation, yielding a reliable categorization into 10 whistle types comprising 7 signature and 3 non-signature whistle types.

\begin{figure}[t]
    \centering
    \includegraphics[width=1\textwidth]{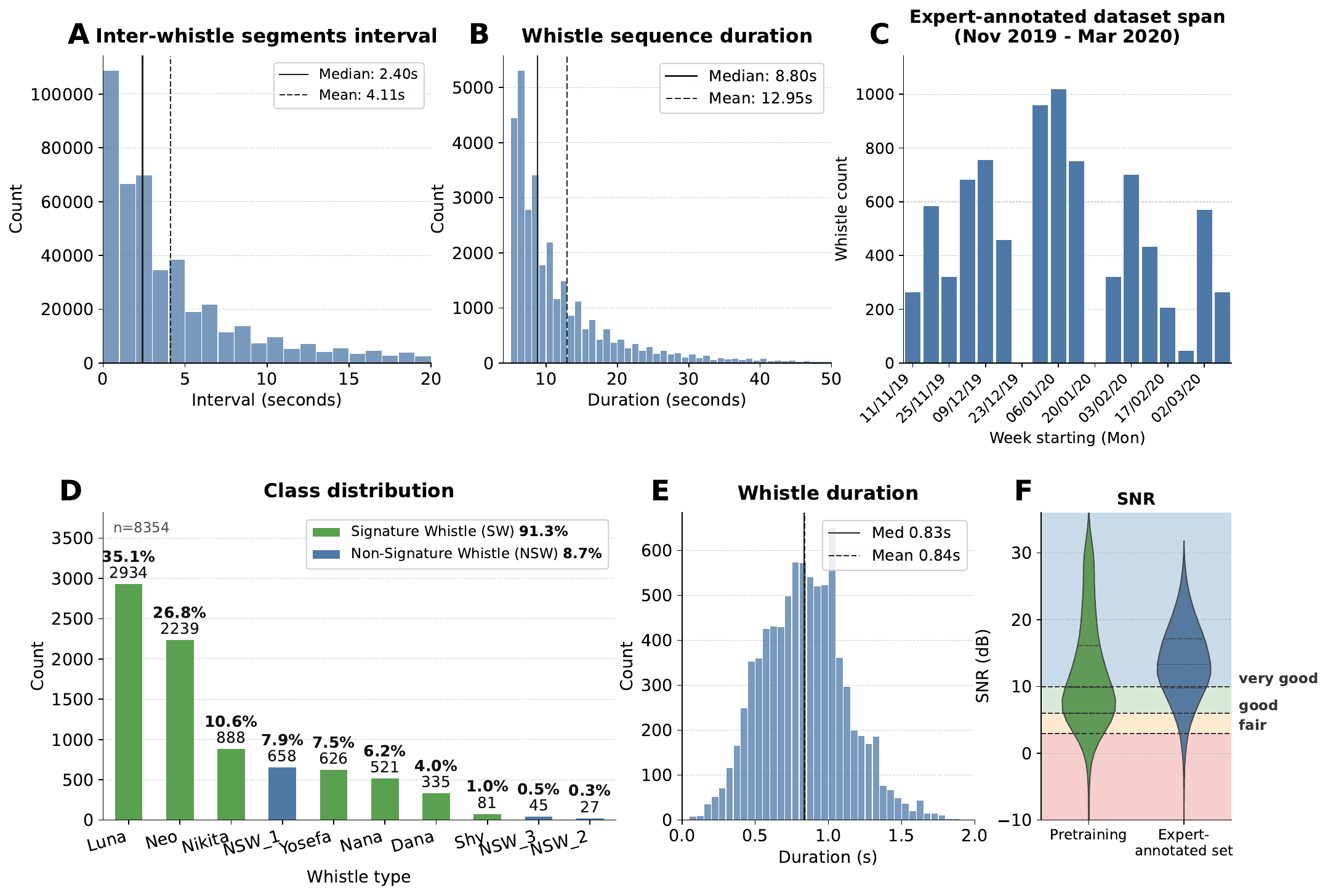}
    \caption{\textbf{Analyses of Whistle Properties in OpenWhistle.} A–B) Temporal structure: distributions of inter-whistle intervals (A) and whistle sequence durations (B).
    C–E) Expert-annotated subset: temporal coverage (C), class distribution (D), and whistle duration (E).
    F) Signal-to-noise ratio (SNR) for the full dataset and the expert-annotated subset.}
    \label{fig:expert_set}
\end{figure}

See Sec.~\ref{app:pipeline_details} for annotation pipeline details. OpenWhistle includes two complementary components: (i) a large-scale pretraining corpus and (ii) a curated expert-annotated dataset.

\subsection{Pretraining Dataset}
\label{sec:data_pretraining_set}

\paragraph{Scale and Coverage.}

The pretraining dataset comprises $\sim$114 hours of raw audio, with an estimated 180,000 whistles across 33,267 sequences. Recordings span over five years (2019--2024), offering longitudinal coverage of 5 identified individuals and enabling analysis of long-term variation, including potential drift in whistle production and social dynamics.

\paragraph{Acoustic Properties.}
Whistle sequences have an average duration of 12.95\,s (SD\,=19.9\,s), ranging from 5 to 246\,s, with a mean interval of 4.11\,s between whistle segments, yielding dense vocal sequences suitable for self-supervised learning. The dataset preserves overlapping vocalizations and environmental sounds, reflecting the realistic acoustic conditions in which the dataset was recorded.

\subsection{Expert-annotated Set}
\label{sec:data_expert_set}

\paragraph{Composition}

Using our annotation pipeline, 8,354 whistles were categorized into 10 categories: 7,624 (91.3\%) signature whistles across 7 types and 730 (8.7\%) non-signature whistles across 3 types, serving as ground truth for downstream tasks. The distribution is highly imbalanced, reflecting natural production frequencies with a few dominant signature whistles and several rare categories.

\paragraph{Acoustic Properties.}
Whistles in the expert-annotated dataset have a mean duration of 0.84\,s (SD\,=\,0.29\,s, range 0.04--2.21\,s), reflecting substantial variability across categories. Acoustic quality is high, with a median signal-to-noise ratio (SNR) of 13.3\,dB, which is above the full pretraining corpus (9.9\,dB). This reflects a manual curation process that favors clear and minimally overlapping vocalizations. Figure~\ref{fig:expert_set} (D, E, F) summarizes class distribution, temporal variability, and quality metrics.

\section{Benchmark Definition}
\label{sec:benchmark_definition}

\begin{figure}[t]
    \centering
    \includegraphics[width=1\textwidth]{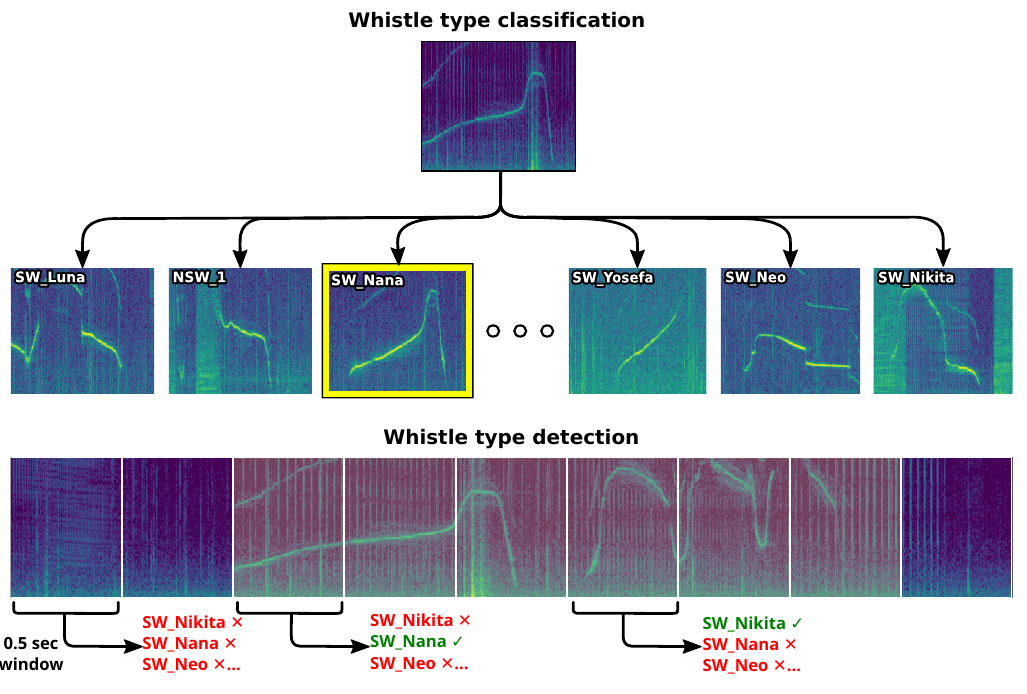}
    \caption{\textbf{Evaluation Tasks.} The two evaluation tasks: (1) whistle type classification, where isolated whistle segments are assigned a predefined category, and (2) whistle-type detection, where fixed 0.5\,s segments are labeled with a category if a whistle is present, or categorized as background otherwise.}
    \label{fig:tasks}
\end{figure}

\subsection{Tasks}
\label{sec:tasks}
We propose two benchmark tasks (classification and detection) grounded in established bioacoustic evaluation practice~\citep{stowell2022computational, hagiwara2023beans}, but adapted to the specific demands of dolphin vocal analysis.

\paragraph{Whistle-Type Classification.}
Given an isolated whistle segment (Figure~\ref{fig:tasks}, top), the model must assign it to one of the 10 whistle types (7 signature, 3 non-signature). The task uses the full expert-annotated set of 8,354 whistles and preserves its natural class imbalance (Figure~\ref{fig:expert_set}D). Performance is reported as macro-averaged F1, so that each type contributes equally regardless of its frequency.

\paragraph{Whistle-Type Detection.}
Given a fixed-length segment drawn from a continuous recording (Figure~\ref{fig:tasks}, bottom), the model must identify which whistle types, if any, are present. Following a standard sliding-window approach, recordings are divided into 0.5\,s segments, each assigned a multi-label prediction over whistle categories (i.e., a binary decision per class, with an all-zero vector for background). The dataset comprises 400 instances per whistle type across 7 classes, balanced with 2,800 background segments. Performance is assessed using mean average precision (mAP)~\citep{hagiwara2023beans}.

Together, these tasks span the core computational pipeline of dolphin communication: from detecting vocal activity in continuous streams to characterizing individual identity and repertoire structure.

\subsection{Evaluation Protocol}
\label{sec:eval_protocol}

All models are evaluated using a linear probing setup with fixed train/validation/test (70\% / 15\% / 15\%). A logistic regression classifier is trained on top of frozen segment representations. Splits are constructed at the session level: all whistles originating from the same recording session are assigned to a single split. This ensures that no acoustic context is shared between training, validation, and test sets, preventing session-level leakage. Despite this constraint, sessions are distributed so that each split approximately preserves the class distribution of the full set.

The regularization parameter $C$ is selected on the validation set. Uncertainty is estimated via bootstrap, repeatedly sampling the test set with replacement ($N=1000$), reporting the full-test score with the standard deviation across resamples.

\section{Experiments}
\label{sec:experiment}

\subsection{Models and Baselines}
\label{sec:exp_models}


We evaluate linear probes on frozen representations from three groups. \emph{Classical acoustic features} include spectral features, MFCCs and the mean spectrogram. \emph{Off-the-shelf pretrained models} include AVES-core and AVES-bio~\cite{Hagiwara:etal:2022}, pretrained on general audio and on animal vocalizations respectively, and BioLingual~\citep{robinson2023transferablemodelsbioacousticshuman}, an audio-text model trained on AnimalSpeak. \emph{Models pretrained on OpenWhistle} use the pretraining corpus of Sec.~\ref{sec:data_pretraining_set} with two self-supervised architectures. wav2vec 2.0~\citep{Baevski:etal:2020} is trained from scratch and was chosen for its discrete latent codebook, which learns a vocabulary of quantized acoustic units directly from raw audio. AVES-bio, a HuBERT-based model, is trained either from scratch or by continual pretraining from the released checkpoint. The from-scratch models, wav2vec 2.0 and AVES-bio, test whether OpenWhistle alone, without any external pretraining data, is sufficient to learn useful representations, while comparing AVES-bio with and without OpenWhistle pretraining measures the contribution of in-domain data within the same architecture. The from-scratch models run natively at 44.1\,kHz, while continual pretraining keeps the 16\,kHz input of the AVES-bio checkpoint and uses multi-band encoding~\cite{sarkar2026} to retain frequencies above 8\,kHz (App.~\ref{app:pretraining}). The linear probes are implemented as logistic regression classifiers trained with the \texttt{lbfgs} solver. We tune the inverse regularization strength over $C \in \{0.1, 1.0, 10.0\}$ on the validation set and set the maximum number of solver iterations to 20,000. 

\subsection{Results}
\label{sec:exp_results}

\begin{table}[h]
  \centering
  \resizebox{\linewidth}{!}{
  \begin{tabular}{l l cc}
    \toprule
    Method & Pretraining & Classification (macro-F1) & Detection (mAP) \\
    \midrule
    Chance level         & --                                & 10.0 & 7.8  \\
    \midrule
    \multicolumn{4}{l}{\textit{Classical acoustic features}} \\
    Spectral features    & --                                & $14.2\pm0.7$ & $26.3\pm1.2$ \\
    MFCCs                & --                                & $21.2\pm1.0$ & $33.9\pm1.8$ \\
    Mean spectrogram     & --                                & $32.1\pm1.3$ & $47.8\pm2.1$ \\
    \midrule
    \multicolumn{4}{l}{\textit{Off-the-shelf pretrained models}} \\
    AVES-core            & General audio (AudioSet, FSD50K)  & $51.6\pm2.3$ & $57.4\pm2.1$ \\
    BioLingual           & Audio-text (AnimalSpeak)          & $54.0\pm2.7$ & $66.5\pm2.2$ \\
    AVES-bio             & Animal vocalizations (AudioSet, VGGSound) & $54.4\pm2.6$ & $65.0\pm2.3$ \\
    \midrule
    \multicolumn{4}{l}{\textit{Pretrained on OpenWhistle (ours)}} \\
    AVES-bio$^{\ddagger}$ & OpenWhistle (from scratch)       & $61.0\pm3.8$ & $76.2\pm2.0$ \\
    wav2vec 2.0           & OpenWhistle (from scratch)        & $62.0\pm2.8$ & $\mathbf{79.6\pm1.8}$ \\
    \midrule
    \multicolumn{4}{l}{\textit{Continual pretraining on OpenWhistle (ours)}} \\
    AVES-bio$^{\dagger}$  & AVES-bio checkpoint $\rightarrow$ OpenWhistle & $\mathbf{64.7\pm2.6}$ & $73.9\pm2.1$ \\
    \bottomrule
  \end{tabular}
  }
    \caption{Performance on whistle-type classification (macro-F1) and whistle-type detection (mAP). Models are grouped by representation and pretraining strategy. All use linear probing with a logistic regression classifier on frozen representations. Scores are reported as percentages. Each entry gives the full-test metric $\pm$ the standard deviation over 1,000 bootstrap resamples of the test set. $^{\dagger}$Continual pretraining at 16\,kHz with multi-band encoding; $^{\ddagger}$trained natively at 44.1\,kHz (App.~\ref{app:pretraining}).}
  \label{tab:results}
\end{table}
Table~\ref{tab:results} compares linear probes across representations and pretraining strategies. We report three complementary findings.

\paragraph{OpenWhistle supports effective self-supervised representation learning.}
The models trained from scratch on OpenWhistle, wav2vec 2.0 and AVES-bio, substantially outperform classical acoustic descriptors (by at least 28.9 macro-F1 for classification and 28.4 mAP for detection over the strongest hand-crafted baseline) and exceed all off-the-shelf pretrained models. This indicates that the dataset is sufficiently large and structurally rich to support self-supervised pretraining directly from raw audio, without relying on transfer from external corpora. Consistent with this, classification performance of wav2vec 2.0 grows with the amount of pretraining data (App.~\ref{app:datasize}). To our knowledge, this is the first application of large-scale self-supervised pretraining directly on dolphin vocalization data.

\paragraph{Pretraining on OpenWhistle drives performance.}
Off-the-shelf pretrained models transfer reasonably well, clearly outperforming classical features, with the strongest off-the-shelf performance at 54.4 macro-F1 (AVES-bio) and 66.5 mAP (BioLingual). Pretraining on OpenWhistle helps further: within the AVES-bio architecture, it improves the off-the-shelf model by 6.6 to 10.3 macro-F1 and 8.9 to 11.2 mAP depending on the initialization. Continual pretraining, which combines the broad pretraining data of the original checkpoint with OpenWhistle, reaches the best classification result (64.7 macro-F1), in line with findings that diverse pretraining data mixes improve bioacoustic encoders~\cite{Miron:etal:2026}. Together with the from-scratch results, this indicates that the benefit of in-domain pretraining is not specific to one architecture.

\paragraph{Both tasks remain unsolved.}
The best models reach 64.7 macro-F1 for classification and 79.6 mAP for detection. While these results demonstrate that the tasks can be learned in practice and benefit from in-domain data, performance remains imperfect, leaving meaningful room for improvement.

This remaining headroom is critical because both tasks underpin downstream analyses of dolphin communication. Reliable detection is required to quantify vocal activity and extract whistle sequences from continuous recordings, forming the basis of any large-scale analysis. Whistle-type classification, in turn, enables the study of signature whistles use and vocal repertoire structure, which are central to understanding social interactions and communication dynamics. Improving performance on these tasks directly expands the scope and reliability of computational analyses of dolphin vocal behavior.

Together, these results position OpenWhistle as both a useful pretraining resource and a challenging benchmark for tracking future progress on these biologically relevant central tasks.

\section{Conclusion}
\label{sec:conclusion}

We introduced OpenWhistle, the largest publicly available dataset of dolphin vocalizations to date. The dataset consists of two complementary components. First, a large-scale corpus of whistles collected over five years from a small pod of known individuals in a natural marine environment. Second, a richly annotated subset of expert-labeled whistles, enabling controlled evaluation of fine-grained tasks. Compared to existing datasets, which are typically small, short-term, or not publicly available, OpenWhistle combines scale, longitudinal continuous coverage, and detailed annotation within a single-species setting, enabling the study of dolphin communication at an unprecedented level of detail. 

We showed that the dataset is sufficiently large and structured to support self-supervised representation learning. Models trained from scratch on OpenWhistle with two different architectures achieve strong performance on both detection and classification, demonstrating that meaningful acoustic features can be learned from raw audio at this scale. Pretraining on OpenWhistle also improves an off-the-shelf bioacoustic model on both tasks, showing that in-domain data drives performance. We further demonstrated that whistle-type classification and detection constitute a challenging benchmark that requires fine-grained, intra-species discrimination. The proposed tasks isolate core computational challenges in dolphin vocal analysis: locating whistle types in continuous streams and discriminating between structurally similar whistle types linked to individual identity. Performance gains from in-domain training, together with remaining errors, indicate that these tasks probe non-trivial acoustic structure rather than superficial cues.

Overall, OpenWhistle enables new directions for studying dolphin communication, including the analysis of vocal sequences, evolution of the vocal repertoire over time, and interaction dynamics as done in \cite{mustun2024whistle}. By releasing the dataset, processing pipeline, and evaluation protocol, we aim to provide a foundation for developing models that capture fine-grained acoustic structure within species.

\section{Future Work}
\label{sec:future_work}

OpenWhistle is part of an ongoing data collection effort. Future releases will expand the dataset with additional audio and extracted whistles, further increasing its scale and temporal coverage. We also plan to extend the expert-annotated subset by labeling more whistles across different periods of the five-year recording span, enabling more robust evaluation and supporting the study of temporal variability and less frequent whistle types. Finally, contextual and video data are available at the site, and future work will explore their integration for multimodal analysis.

\section{Limitations}
\label{sec:limitations}

\paragraph{Geographic and demographic scope.}

All recordings come from a single site and from a pod of five individuals, limiting dataset diversity; results should be validated on independent groups.

\paragraph{Temporal coverage and recording bias.}

Recording coverage is uneven across the dataset, with intermittent sampling within each year, concentration at specific times of day, and a full gap in 2022 (Figure~\ref{fig:long_data}A–B). The dataset also spans from late 2019 to early 2024, with variable recording density across periods. As a result, the data does not provide uniform temporal sampling of dolphin vocal activity, and models may reflect the conditions and behaviors most represented in the corpus.

\paragraph{Pipeline recall gaps.}

The detection CNN achieves a recall of 97.99\%, implying that an estimated $\sim$3,700 whistles are not captured in the dataset. On the CNN test set, missed whistles had a much lower median SNR than detected ones (2.1 vs.\ 11.5\,dB; $n=168$ and $8{,}186$; $p=1.2\times10^{-61}$), so low-SNR vocalizations are under-represented and the absence of a whistle type in the corpus does not imply it was not produced.

\paragraph{Limited temporal coverage of expert annotations.}

The expert-labeled dataset spans only a short period (5 months) within the five-year recording window; as a result, model performance measured on this subset may not generalize to the entire dataset.

\paragraph{Class imbalance.}

The annotated dataset is highly imbalanced (Figure~\ref{fig:expert_set}D), with some categories having fewer than 100 examples. Although the benchmark includes all 10 types and uses macro-F1, the rarest type has as few as 4 test whistles, making scores on rare types noisy, and imbalance may bias models toward frequent categories; the balanced configuration (App.~\ref{app:whistle_classification}) provides a complementary evaluation on the six best-represented types.

\paragraph{Scope of the benchmark.}
OpenWhistle evaluates models on whistle-type detection and classification, not on semantic interpretation or communicative meaning. The proposed tasks are intended as foundational steps for large-scale computational analyses of dolphin vocal behavior, including vocal activity, repertoire structure, individual identity, and temporal variation. However, strong performance on these benchmarks should not be interpreted as evidence that a model has inferred the meaning or communicative function of dolphin whistles. Future work will require additional behavioral, social, and contextual annotations to evaluate models on questions related to signal function and meaning.

\section{Ethics and Broader Impact}
\label{sec:ethics_impact}

All recordings were collected in a semi-natural environment without interfering with dolphin behavior. No animals were trained or constrained in any form. Acoustic recording was passive, using fixed and hidden hydrophones not altering the animals’ environment. Interactions between dolphins and humans were only initiated by dolphins. The dataset contains no human subjects and follows standard passive acoustic monitoring practices. OpenWhistle is released under CC-BY 4.0, and the code under the MIT license, to support research in bioacoustics and machine learning. Misuse risks are limited. It enables large-scale study of dolphin communication, including structure, non-invasive monitoring, and conservation, and provides a benchmark for fine-grained acoustic modeling.


\section*{Acknowledgments}
\label{sec:acknowledgments}

We extend our sincere gratitude to the dedicated individuals who contributed to the success of this research endeavor. This project was supported by the FIRE PhD program, funded by the Bettencourt Schueller foundation, and the EURIP graduate program (ANR-17-EURE-0012), and PSL-Neuro, funded by PSL University to F. Mustun, the European Union's Horizon 2020 research and innovation program under the Marie Skłodowska-Curie grant agreement No 945304 – Cofund AI4theSciences, and Labex MemoLIFE to C. Semenzin, International Emerging Actions (IEA CNRS, IEA00637), and International Research Project (IRP CNRS) to G. Sumbre and G. de Polavieja. G. de Polavieja was also supported by the Champalimaud Foundation.

\bibliographystyle{plainnat}
\bibliography{bibliography/chiara, bibliography/faadil, bibliography/roberto, bibliography/yair, bibliography/ema, bibliography/pablo}

\newpage
\appendix

\section{Additional Data Collection Information}

\subsection{Equipment and Recording Protocol} 
\label{app:setup}

Acoustic recordings were obtained using three Brüel \& Kjær\textsuperscript{\textregistered} 8104 hydrophones connected to 1704 preamplifiers and a National Instruments\textsuperscript{\textregistered} PCI-4474 acquisition card, sampling at 96\,kHz. Recordings were conducted daily for an average of 11.7 hours at varying times of day. Data acquisition was automated using scheduled \texttt{crontab} commands using a Linux HP Z400 computer. The recording period spans from 12 November 2019 to 28 March 2024, totaling 7,495 recording sessions and 6,271.78 hours of usable audio (Figure~\ref{fig:long_data}A).

\subsection{Dolphins and Individual Metadata}
\label{app:dolphins}

\begin{figure}[h]
    \centering
    \includegraphics[width=1\textwidth]{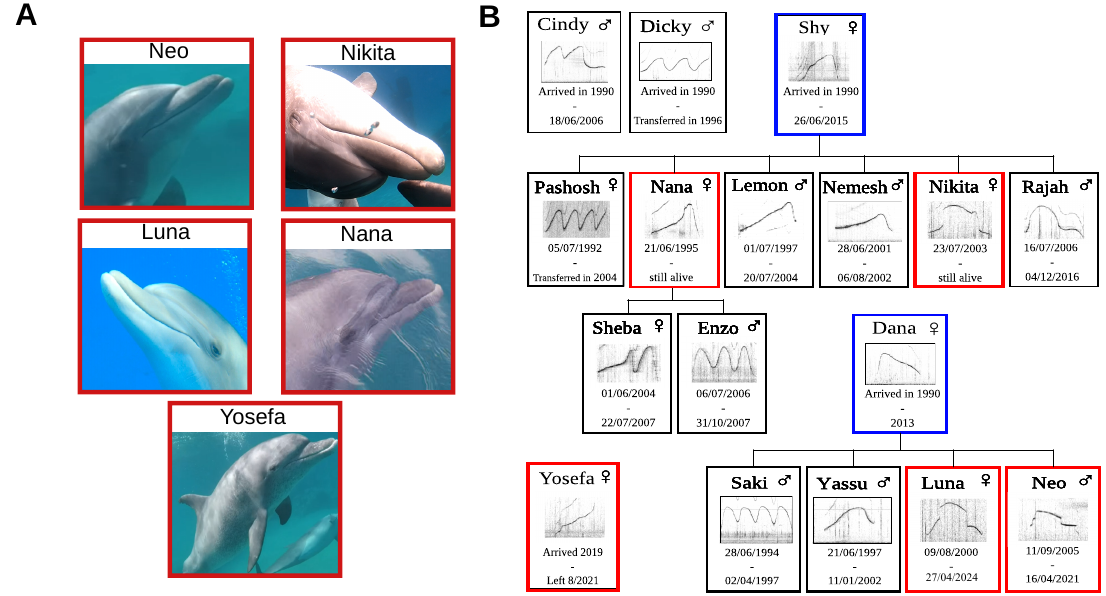}
    \caption{A) Photographs of the five dolphins present during the recording period. B) Family tree of the pod, indicating sex and signature whistles (SW). Dolphins present during the recording period are highlighted in red. Dolphins not present but whose signature whistles appear in the dataset are shown in blue.}
    \label{fig:dolphins}
\end{figure}

At the beginning of the recording period, the pod comprised five dolphins (Figure~\ref{fig:dolphins}A): Luna (female, 20 years), Nana (female, 25 years), Nikita (female, 17 years), and Neo (male, 15 years), all belonging to Tursiops truncatus ponticus and forming a stable social group with well-documented family relationships. In addition, a solitary Tursiops aduncus female, Yosefa, visited the group intermittently, introducing an external social component. Her signature whistle was identified using the SIGID (Signature Identification) procedure~\cite{janik_identifying_2013}.

For all resident individuals at Dolphin Reef, we have associated metadata, including identity, familial relationships, and their corresponding signature whistles (Figure~\ref{fig:dolphins}B). This enables linking acoustic signals to known individuals and supports analyses of vocal identity and social structure~\cite{mustun2024whistle}.

\section{Dataset Construction Pipeline}
\label{app:pipeline_details}

\begin{figure}[h]
    \centering
    \includegraphics[width=1\textwidth]{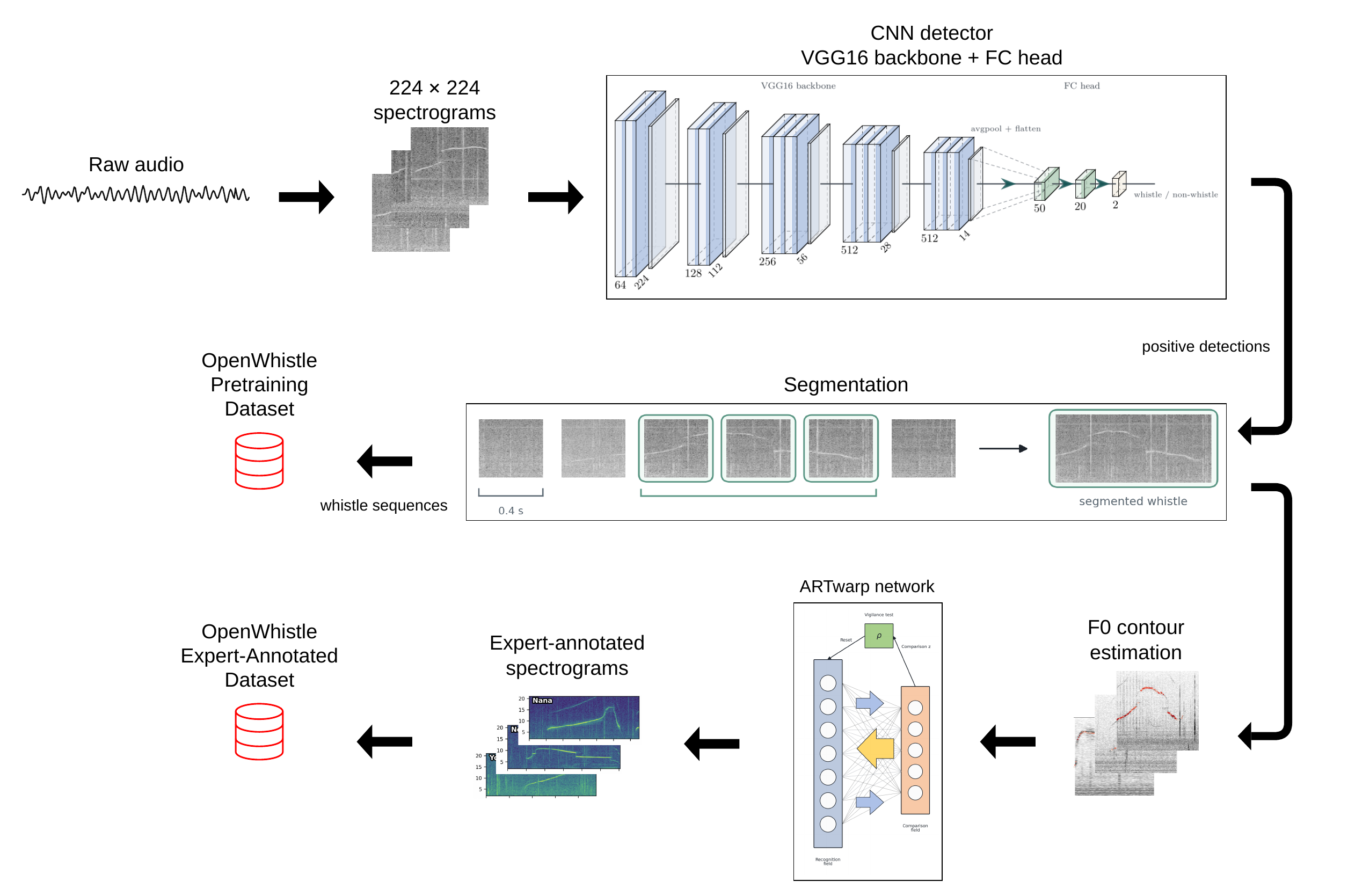}
    \caption{Dataset construction pipeline. Raw audio is converted to 224$\times$224 spectrograms and processed by a VGG16-based detector. Positive detections are segmented and then used in two branches: one branch groups detections into whistle sequences to construct the pretraining corpus, while the other applies F0 estimation, ARTwarp categorization, and expert annotation to construct the expert-annotated dataset.}
    \label{fig:pipeline}
\end{figure}

\paragraph{Overview.}
Figure~\ref{fig:pipeline} summarizes the dataset construction pipeline. Raw audio recordings are first generated into fixed-duration spectrogram windows and processed with a VGG16-based binary detector to identify whistle-containing windows. Positive detections are then segmented and grouped into whistle sequences to construct the large-scale pretraining corpus. For the expert-annotation branch, segmented whistles are processed to estimate fundamental-frequency (F0) contours. Following the procedure introduced in~\citet{mustun2024whistle}, these contours are categorized with ARTwarp to obtain initial whistle-category assignments, which are manually reviewed and corrected from spectrogram visualizations to produce the expert-annotated dataset.

\paragraph{Whistle detection.}
Whistle detection is performed with a binary spectrogram classifier based on a VGG16 backbone~\cite{simonyan2015vgg} initialized from ImageNet-pretrained weights~\cite{deng2009imagenet}. Audio data is split into non-overlapping 0.4\,s windows. Each window is converted to a log-power spectrogram using a 1024-sample Blackman window, an FFT size of 1024, and a hop size of 512 samples. Spectrograms are cropped to 2--22\,kHz, min--max normalized, resize to 224$\times$224 pixels, replicate across three channels, and normalize using ImageNet statistics.

Following~\cite{korkmaz2023}, the original VGG16 classifier was replaced with a lightweight fully connected head with hidden dimensions 50 and 20. The full network was fine-tuned for whistle-versus-noise classification using cross-entropy loss and Adam with learning rate $10^{-5}$, mini-batches of size 4, early stopping, and a ReduceLROnPlateau scheduler. Training and evaluation used balanced whistle/noise windows with session-disjoint splits; the final dataset contained 53,828 training, 5,980 validation, and 16,708 test windows. For sequence-level summaries, positive windows were grouped using a maximum inter-detection gap of 6\,s, retaining sequences between 2 and 20\,s.

As external robustness checks, the trained detector achieve F1 scores of 0.904 on a WMMSD binary clip benchmark~\cite{sayigh2016watkins}, 0.907 on a broader WMMSD delphinid-versus-clear-noise proxy benchmark, and 0.966 on a DCLDE proxy subset~\cite{roch2025dclde,roch_automated_2011}, without retraining.

\paragraph{F0 extraction and ARTwarp categorization.}

Fundamental-frequency (F0) contours are estimated with a dolphin-specific CREPE model~\cite{kim2018crepe,best_f0_2025}. Because dolphin whistles extend above the pitch range targeted by the original CREPE model, we use the frequency-compression procedure from~\cite{best_f0_2025}: audio is processed with \texttt{compress=20}, and decoded F0 estimates are multiplied back by the same factor. F0 is estimated every 5\,ms using the \texttt{weighted\_argmax} decoder. Contours with fewer than 5\% of frames above a confidence threshold of 0.05 are flagged as low-confidence.

For downstream whistle-type classification, the extracted contours were categorized with ARTwarp~\cite{deecke2006automated}, which combines dynamic time warping with an adaptive resonance theory network. The vigilance parameter was set to 90, following~\cite{deecke2006automated}; all other ARTwarp parameters were kept at their default values.

\paragraph{Expert review and annotation.}

ARTwarp assignments were treated as proposals rather than final labels. An expert annotator inspected the spectrogram and F0 contour of each whistle alongside its ARTwarp assignment and the template contours of the 10 whistle types, then confirmed or corrected the assignment. A small number of whistles were discarded because they were too faint, too short, or did not match any of the templates.
To assess the consistency of this review, two annotators independently re-labeled 100 whistles sampled from the expert-annotated subset in proportion to its whistle-type distribution, without access to the ARTwarp assignments. On the 84 whistles both judged classifiable, they assigned the same type in 82.1\% of cases (Cohen's $\kappa$~\cite{cohen1960}: 0.76, 95\% CI 0.65--0.86; Gwet's AC1~\cite{gwet2008}: 0.81), indicating substantial agreement~\cite{landis1977}. Including whistles judged \textit{uncertain/unusable} by either annotator, agreement drops to 73.0\% ($\kappa=0.66$, AC1 $=0.71$). Given the small sample, dominated by the most frequent types, these figures provide an initial estimate of annotation reliability.

\section{Pretraining Setup}
\label{app:pretraining}

\paragraph{Wav2vec 2.0.}
We pretrain a wav2vec 2.0 model \cite{Baevski:etal:2020} on the OpenWhistle corpus following a standard self-supervised setup. Training is conducted for 400k steps on 16 V100 GPUs, with a maximum audio budget of 882,000 samples (20 seconds) per GPU and a gradient accumulation factor of 1, yielding a maximum global audio budget of 320 seconds per update. The number of audio segments per update varies with their duration. Optimization uses AdamW~\cite{Loshchilov:Hutter:2018} with $\beta_1=0.9$, $\beta_2=0.98$, $\epsilon=10^{-6}$, a peak learning rate of $5\times10^{-4}$, reached after 32k warmup steps and followed by linear decay, and weight decay of 0.01. Mixed precision is used to improve efficiency~\cite{micikevicius2017mixed}. The quantization module employs two codebooks of size 320, trained with a Gumbel-softmax temperature schedule following $\tau_t=\max\{2.0\times0.999995^{t},\,0.5\}$, where $t$ denotes the number of optimizer updates~\cite{Jang:etal:2017, Maddison:etal:2017}.

To account for the higher sampling rate of 44.1 kHz compared to the 16 kHz setting of speech benchmarks, we adapt the feature encoder to preserve the relative temporal resolution of the original architecture, following the approach introduced in \cite{dolph2vec}. The feature encoder has a total stride of 960 samples, corresponding to approximately 21.77~ms at 44.1~kHz. The Transformer comprises 12 layers, a hidden dimension of 768, a feed-forward dimension of 3,072, eight attention heads, and pre-layer normalization.

\paragraph{AVES-bio.}
We use the HuBERT base architecture underlying AVES-bio~\cite{Hagiwara:etal:2022} (12 transformer layers, 768 hidden units) under two pretraining regimes. For training from scratch on OpenWhistle, we follow the original two-stage AVES schedule: first-stage targets are obtained by k-means clustering of MFCC features, and second-stage targets by k-means clustering of the 6th-layer features of the first-stage model, with 200 clusters at both stages. Each stage is trained for 100k optimizer updates with a peak learning rate of $2.0\times10^{-4}$.
For continual pretraining, we initialize the model from the released AVES-bio checkpoint and perform a single additional masked-prediction training phase. Targets are generated on the multiband OpenWhistle training data by fitting a new 200-cluster k-means model to the 6th-layer representations of the released AVES-bio encoder. Continual pretraining runs for 50k optimizer updates with a peak learning rate of $5.0\times10^{-5}$; the checkpoint with the lowest validation loss, obtained at 45k updates, is used for evaluation.

The two regimes require different treatments of the sampling rate. For training from scratch, we train the model natively at 44.1\,kHz while retaining the original feature encoder, whose total stride of 320 samples yields a frame hop of approximately 7.3\,ms instead of 20\,ms at 16\,kHz. For continual pretraining, the released AVES-bio checkpoint operates at 16\,kHz, which would discard all frequencies above 8\,kHz, a range in which dolphin whistles commonly lie. We therefore use the multi-band encoding method of Sarkar et al.~\cite{sarkar2026}, which represents the spectral content of the 44.1\,kHz recordings through baseband-compatible sub-bands.

\section{Pretraining Data-Size Ablation}
\label{app:datasize}

To assess how much the scale of OpenWhistle contributes to representation quality, we pretrained wav2vec 2.0 from scratch on 10\%, 50\% and 100\% of the pretraining corpus and evaluated each model on whistle-type classification with the linear-probing protocol of Sec.~\ref{sec:eval_protocol}. For classification, we report macro-F1 on the \texttt{unbalanced} configuration, as in Table~\ref{tab:results}, and on the \texttt{balanced} configuration (App.~\ref{app:whistle_classification}). The 10\% and 50\% subsets were drawn by stratified random sampling with a fixed seed, stratifying on recording year, hydrophone channel and clip duration. Each subset therefore preserves the temporal, spatial and duration structure of the full corpus, so the ablation varies the quantity of data rather than its coverage.

\begin{table}[h]
  \centering
  \resizebox{\linewidth}{!}{
  \begin{tabular}{l c c c c c c}
    \toprule
    Data-size & Train (h) & Validation (h) & Total (h) & Unbalanced & Balanced & Detection\\
    \midrule
    10\%  & 10.16 & 1.18 & 11.34 & $20.6\pm1.0$ & $33.5\pm2.1$ & $20.5\pm1.4$ \\ 
    50\%  & 51.23 & 6.06 & 57.29 & $55.0\pm1.6$ & $77.5\pm1.9$ & $77.5\pm2.0$ \\
    100\% & 102.17 & 12.11 & 114.28 & $\mathbf{62.0\pm2.8}$ & $\mathbf{81.4\pm1.8}$ & $\mathbf{79.6\pm1.8}$ \\
    \bottomrule
  \end{tabular}}
  \caption{Effect of pretraining data size. Wav2vec 2.0 is pretrained from scratch on stratified subsets of the OpenWhistle pretraining corpus and evaluated by linear probing on whistle-type classification (macro-F1 on the \texttt{unbalanced} and \texttt{balanced} configurations) and whistle-type detection (mAP). Each entry gives the full-test metric $\pm$ the standard deviation over 1,000 bootstrap resamples of the test set.}
  \label{tab:datasize}
\end{table}

Performance increases monotonically with the amount of pretraining data on all three evaluations (Table~\ref{tab:datasize}). With 10\% of the corpus, the model performs poorly: on the \texttt{unbalanced} configuration it falls between spectral features (14.2) and MFCCs (21.2), and on detection it is below all classical features (Table~\ref{tab:results}), consistent with self-supervised pretraining requiring substantial data to learn useful representations. Most of the gain is reached with 50\% of the corpus, particularly for detection (77.5 vs.\ 79.6 mAP with the full corpus). Going from 50\% to 100\% of the corpus still adds 7.0 macro-F1 on the \texttt{unbalanced} configuration, the largest gain of the three evaluations, indicating that performance on the full 10-class task has not saturated at the current scale. Future releases of OpenWhistle, which will add recordings and temporal coverage (Sec.~\ref{sec:future_work}), may therefore yield further gains.

All budgets use the architecture and optimization hyperparameters of the full-corpus model (App.~\ref{app:pretraining}). Smaller budgets may benefit from budget-specific tuning, so the low-data results are likely conservative.

\section{Additional Analysis of Whistle-Type Classification}
\label{app:whistle_classification}

OpenWhistle provides two configurations of the classification set. The \texttt{unbalanced} configuration, used as the official benchmark (Sec.~\ref{sec:tasks}), contains all 8,354 expert-annotated whistles across the 10 types with their natural class distribution. The \texttt{balanced} configuration contains 3,000 whistles from the six best-represented types (NSW\_1, SW\_Luna, SW\_Nana, SW\_Neo, SW\_Nikita, SW\_Yosefa), 500 per type, subsampled from \texttt{unbalanced} within the same session-level splits. It removes the effect of class imbalance and of the rare types, and suits comparisons focused on the frequent types.

\begin{table}[h]
  \centering
  \begin{tabular}{l cc}
    \toprule
    Method & \texttt{unbalanced} (10 types) & \texttt{balanced} (6 types) \\
    \midrule
    Chance level      & 10.0 & 16.7 \\
    \midrule
    \multicolumn{3}{l}{\textit{Classical acoustic features}} \\
    Spectral features & $14.2\pm0.7$ & $33.5\pm2.1$ \\
    MFCCs             & $21.2\pm1.0$ & $44.2\pm2.2$ \\
    Mean spectrogram  & $32.0\pm1.3$ & $55.0\pm2.2$ \\
    \midrule
    \multicolumn{3}{l}{\textit{Off-the-shelf pretrained models}} \\
    AVES-core         & $51.6\pm2.3$ & $67.6\pm2.1$ \\
    BioLingual        & $54.0\pm2.7$ & $71.0\pm2.1$ \\
    AVES-bio          & $54.5\pm2.6$ & $75.1\pm2.0$ \\
    \midrule
    \multicolumn{3}{l}{\textit{Pretrained on OpenWhistle (ours)}} \\
    AVES-bio & $61.0\pm3.8$ & $80.0\pm1.8$ \\
    wav2vec 2.0 & $62.0\pm2.8$ & $\mathbf{81.4\pm1.8}$ \\
    \midrule
    \multicolumn{3}{l}{\textit{Continual pretraining on OpenWhistle (ours)}} \\
    AVES-bio & $\mathbf{64.8\pm2.6}$ & $75.2\pm2.0$ \\
    \bottomrule
  \end{tabular}
  \caption{Whistle-type classification (macro-F1) on the two configurations of the classification set, with the linear-probing protocol of Sec.~\ref{sec:eval_protocol}. Each entry gives the full-test metric $\pm$ the standard deviation over 1,000 bootstrap resamples of the test set.}
  \label{tab:balanced}
\end{table}

On both configurations, classical features remain below off-the-shelf pretrained models, which remain below the models pretrained from scratch on OpenWhistle (Table~\ref{tab:balanced}). The \texttt{unbalanced} configuration separates these groups more clearly: classical features lose 42 to 58\% of their \texttt{balanced} score, whereas pretrained models lose 14 to 28\%. Continual pretraining behaves differently across configurations: it gives the best \texttt{unbalanced} result (64.7) but performs on par with the off-the-shelf AVES-bio checkpoint on \texttt{balanced} (75.2 vs.\ 75.1), so its gain is concentrated in the configuration that adds the rare types and the natural class imbalance.

\begin{figure}[H]
    \centering
    \includegraphics[width=1\textwidth]{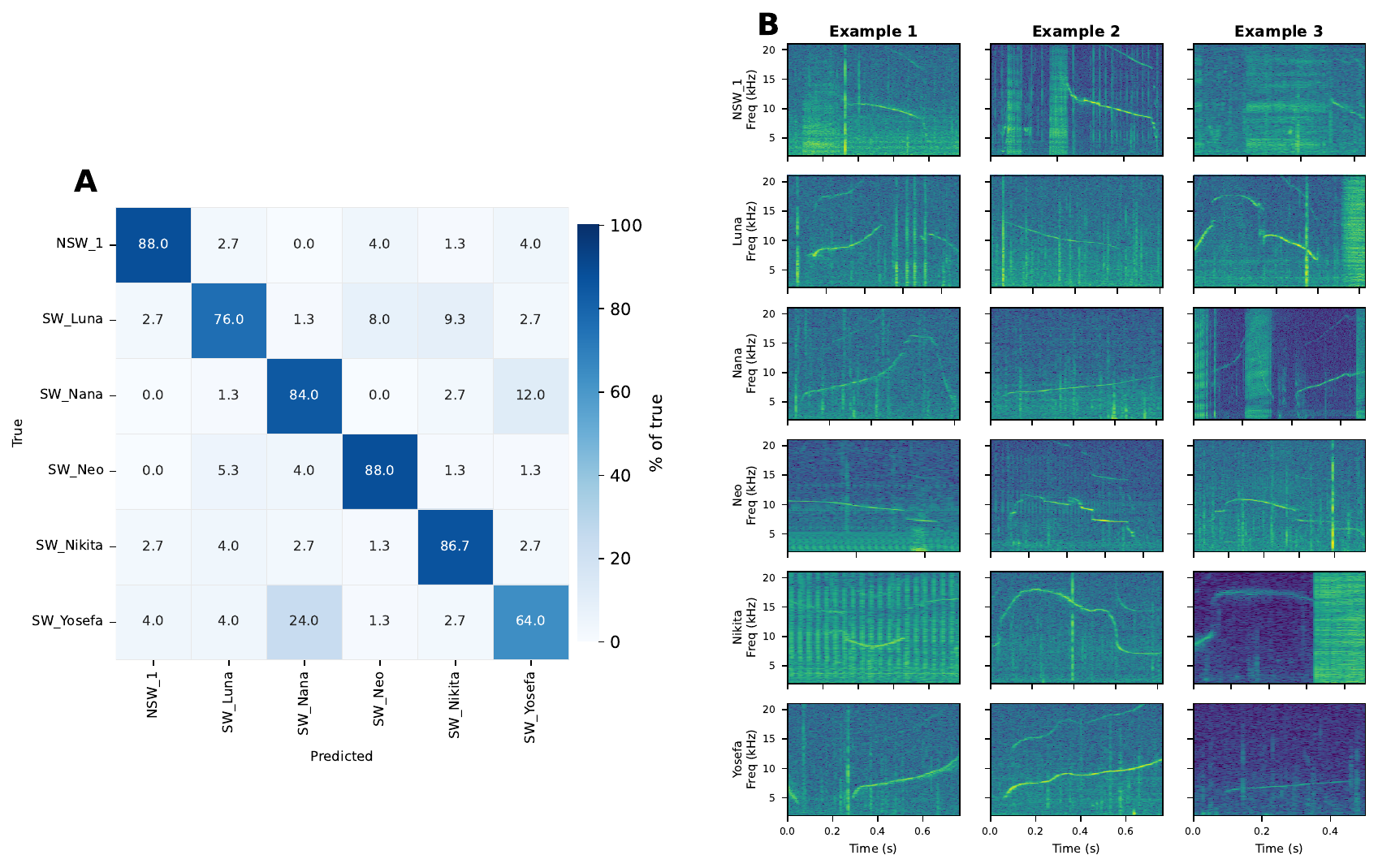}
    \caption{A) Confusion matrix of the wav2vec 2.0 model pretrained on OpenWhistle for whistle-type classification on the \texttt{balanced} configuration (row-normalized, in \%).
    B) Three example spectrograms for each of the six whistle types in the \texttt{balanced} configuration.}
    \label{fig:conf_mat_whistles}
\end{figure}

Because each type has the same number of test whistles in the \texttt{balanced} configuration, we use it to compare per-class errors of the wav2vec 2.0 model (Figure~\ref{fig:conf_mat_whistles}). The dominant diagonal of the confusion matrix shows strong overall performance, but the errors are structured rather than uniform. The most frequent confusion is between the signature whistles of Nana and Yosefa: nearly a quarter of SW\_Yosefa whistles are predicted as SW\_Nana, and the spectrogram examples (Figure~\ref{fig:conf_mat_whistles}B) show that these two signature whistles share similar frequency contours. SW\_Luna is mostly confused with SW\_Nikita and SW\_Neo. The Luna--Nikita pair also accounted for nearly half of the whistle-type disagreements between the two annotators in our agreement study (App.~\ref{app:pipeline_details}), suggesting that some of these errors reflect genuine ambiguity between contours rather than model failure alone. The examples also highlight intra-class variability, with noticeable variations in frequency modulation within the same whistle type. Both inter-class similarity and intra-class variability contribute to the remaining errors, indicating that the task requires fine-grained discrimination of subtle contour differences.


\section{End-to-End Fine-Tuning}
\label{app:finetuning}

To verify that our conclusions do not depend on the linear-probing protocol, we fine-tuned all pretrained models end-to-end, unfreezing the encoder and training a linear head with dropout ($p=0.1$) on each benchmark task. We used AdamW with weight decay $0.01$ and learning rates of $10^{-5}$ for the encoder and $10^{-4}$ for the head, except for BioLingual classification, where the encoder learning rate was $5 \times 10^{-6}$. Learning rates followed a linear schedule with 10\% warmup. Training ran for at most 50 epochs for classification and 30 for detection. We selected the epoch maximizing validation macro-F1 for classification or validation mAP for detection, with early-stopping patience of 15 epochs for classification and 5 epochs for detection (10 for BioLingual and 15 for wav2vec 2.0). After selection, we restarted from the original pretrained encoder and a newly initialized head, reset the optimizer and scheduler, and trained on the combined training and validation sets for the selected number of epochs before evaluating once on the test set. The learning-rate schedule retained its original maximum-epoch horizon during this final training phase. All runs used seed 42.

\begin{table}[t]
    \centering
    \resizebox{\linewidth}{!}{
    \begin{tabular}{@{}llccc@{}}
        \toprule
        \multirow{2}{*}{\textbf{Method}}
          & \multirow{2}{*}{\textbf{Pretraining}}
          & \multicolumn{2}{c}{\textbf{Classification}}
          & \multirow{2}{*}{\textbf{Detection}} \\
        \cmidrule(lr){3-4}
          & & Unbalanced & Balanced & \\
        \midrule
        
        \multicolumn{5}{@{}l}{\textit{Off-the-shelf pretrained models}} \\
        AVES-core
          & General audio (AudioSet, FSD50K)
          & $58.4\pm2.2$ & $78.9\pm1.9$ & $77.5\pm2.8$ \\
        BioLingual
          & Audio-text (AnimalSpeak)
          & $57.3\pm2.0$ & $80.4\pm1.8$ & $73.5\pm1.9$ \\
        AVES-bio
          & Animal vocalizations (AudioSet, VGGSound)
          & $61.4\pm2.6$ & $80.1\pm1.8$ & $76.9\pm2.2$ \\
        \midrule
        \multicolumn{5}{@{}l}{\textit{Pretrained on OpenWhistle (ours)}} \\
        AVES-bio
          & OpenWhistle (from scratch)
          & $\mathbf{64.8\pm3.5}$ & $\mathbf{85.6\pm1.6}$ & $\mathbf{81.4\pm2.0}$ \\
        wav2vec 2.0
          & OpenWhistle (from scratch)
          & $55.8\pm1.9$ & $84.1\pm1.6$ & $81.3\pm2.0$ \\
        \midrule
        \multicolumn{5}{@{}l}{\textit{Continual pretraining on OpenWhistle (ours)}} \\
        AVES-bio
          & AVES-bio checkpoint $\rightarrow$ OpenWhistle
          & $62.2\pm2.6$ & $82.2\pm1.7$ & $78.4\pm1.9$ \\
        \bottomrule
    \end{tabular}%
    }
    \caption{Performance after end-to-end fine-tuning, using the same splits and metrics as Table~\ref{tab:results}. Each entry gives the full-test metric $\pm$ the standard deviation over 1,000 bootstrap resamples of the test set.}
    \label{tab:finetuning}
\end{table}

After fine-tuning, AVES-bio pretrained from scratch on OpenWhistle emerges as the best model on all three evaluations (Table~\ref{tab:finetuning}). Within the AVES-bio architecture, pretraining on OpenWhistle from scratch still improves over the off-the-shelf checkpoint, by 3.4 and 5.5 macro-F1 on the \texttt{unbalanced} and \texttt{balanced} configurations and 4.5 mAP on detection, although these margins are smaller than under linear probing. On \texttt{balanced} and detection, every model pretrained on OpenWhistle outperforms every off-the-shelf model. Fine-tuning improves every model on these two evaluations, with the largest gains for off-the-shelf models (up to 20.1 mAP for AVES-core on detection). 

On \texttt{unbalanced}, however, fine-tuning lowers the scores of wav2vec 2.0 (from 62.0 to 55.8) and of continual pretraining (from 64.7 to 62.2). This is explained by both models largely failing on the rarest types, which have very few training whistles; wav2vec 2.0, for instance, never predicts NSW\_2 or NSW\_3. This is consistent with the observation that fine-tuning can underperform linear probing when few training examples per class are available~\cite{Kornblith:etal:2019,Kumar:etal:2022}. A class-weighted loss mitigates the drop (62.8 and 62.7), but also raises the off-the-shelf AVES-bio to 64.6, leaving most fine-tuned models within one standard deviation of each other on this configuration. With only 4 to 12 test whistles for the rarest types, the \texttt{unbalanced} configuration does not reliably separate fine-tuned models.

\end{document}